%% file: TrolesSchmidMTGenderBias.tex
\documentclass[runningheads]{llncs}
\usepackage{graphicx}
\usepackage{lscape} 
\usepackage{tabularx}
\usepackage{booktabs}
\usepackage{makecell}

\usepackage{csquotes}
\usepackage[hyphens]{url}
\usepackage[official]{eurosym}

\usepackage{graphicx, tikz, pgfplots}
\usepackage{rotating} 
\pgfplotsset{compat=1.17} 

\usepackage{pdfpages}
\usepackage{xcolor}
\definecolor{gray1}{gray}{0.92}
\definecolor{darkgreen}{rgb}{0,0.5,0}
\definecolor{urlLinkColor}{rgb}{0,0,0.5}
\definecolor{LinkColor}{rgb}{0,0,0}
\definecolor{ListingBackground}{rgb}{0.85,0.85,0.85}

\usepackage{color}
\definecolor{LinkColor}{rgb}{0.1,0.1,0.1}
\definecolor{ListingBackground}{rgb}{0.98,0.98,0.98}
\definecolor{gray}{rgb}{0.4,0.4,0.4}
\definecolor{darkblue}{rgb}{0.0,0.0,0.6}
\definecolor{cyan}{rgb}{0.0,0.6,0.6}

\definecolor{snsBlue}{RGB}{1,115,178}
\definecolor{snsLightBlue}{RGB}{86,180,233}
\definecolor{snsOrange}{RGB}{213,94,0}
\definecolor{snsLightOrange}{RGB}{222,143,5}
\definecolor{snsBrown}{RGB}{202,145,97}
\definecolor{snsYellow}{RGB}{236,225,51}
\definecolor{snsGreen}{RGB}{2,158,115}
\definecolor{snsPink}{RGB}{204,120,188}
\definecolor{snsLightPink}{RGB}{251,175,228}
\definecolor{snsGrey}{RGB}{148,148,148}

\usepackage{moreverb}

\usepackage{listings}
\usepackage{caption}
\DeclareCaptionFont{white}{\color{white}}
\DeclareCaptionFormat{listing}{\colorbox[cmyk]{0.43, 0.35, 0.35,0.01}{\parbox{\textwidth}{\hspace{15pt}#1#2#3}}}

\usepackage[ruled,vlined]{algorithm2e}
\DontPrintSemicolon 

\usepackage{amssymb}
\usepackage{amsmath}

\usepackage{subcaption}

\usepackage{wrapfig}
\usepackage{lscape} 
\usepackage{tabularx}
\usepackage{booktabs}
\usepackage{makecell}

\begin{document}
\title{Extending Challenge Sets to Uncover Gender Bias in Machine Translation}
\subtitle{Impact of Stereotypical Verbs and Adjectives}

%
%
\author{Jonas-Dario Troles \and
Ute Schmid}
\authorrunning{Jonas Troles and Ute Schmid}
%
\institute{
\email{jonas.troles@uni-bamberg.de} \\
\email{ute.schmid@uni-bamberg.de}
}
\maketitle              
\begin{abstract}
Human gender bias is reflected in language and text production. Because state-of-the-art machine translation (MT) systems are trained on large corpora of text, mostly generated by humans, gender bias can also be found in  MT. For instance when occupations are translated from a language like English, which mostly uses gender neutral words, to a language like German, which mostly uses a feminine and a masculine version for an occupation, a decision must be made by the MT System. 
Recent research showed that MT systems are biased towards stereotypical translation of occupations.
In 2019 the first, and so far only, challenge set, explicitly designed to measure the extent of gender bias in MT systems has been published. In this set measurement of gender bias is solely based on the translation of occupations. In this paper we present an extension of this challenge set, called \emph{WiBeMT}, with gender-biased adjectives and adds sentences with gender-biased verbs. The resulting challenge set consists of over $70,000$ sentences and has been translated with three commercial MT systems: DeepL Translator, Microsoft Translator, and Google Translate. Results show a gender bias for all three MT systems. This gender bias is to a great extent significantly influenced by adjectives and to a lesser extent by verbs.

\end{abstract}

\input{Introduction}
\input{Methods}
\input{Results}
\input{Discussion}
%
%
%

%
%
%
\bibliographystyle{splncs04}
\bibliography{bib/libraryCorrected}

\end{document}

%% file: Introduction.tex
\section{Introduction}


The problem of unfair and biased models has been recognized as an important problem for many applications of machine learning \cite{Mehrabi2019}. The source of unfairness typically is based on a biases in the training data. In many domains, unfairness is caused by sampling biases. In machine translation (MT), however, the main source of unfairness is due to historical or social biases when there is a misalignment between the world as it is and the values or objectives to be encoded and propagated in a model \cite{Suresh2019,Bolukbasi2016}. One source of imbalance in natural language is the association of specific occupations with gender. Typically, occupations in more technical domains as well as occupations with high social status are associated with male gender \cite{Cheryan2017}. 

In natural language processing (NLP), gender bias has been investigated for word embeddings \cite{Bolukbasi2016}. In analogy puzzles such as \enquote{man is to king as woman is to $x$} generated with \emph{word2vec}\footnote{\url{https://code.google.com/archive/p/word2vec/}} yields $x=queen$ while for \enquote{man is to computer programmer as woman is to $x$}, the output is $x=homemaker$. Only few publications exist that directly address gender bias and MT. Although gender bias is particularly relevant for translations into gender-inflected languages, for instance from English to German, when biased models can result in translation errors \cite{Saunders2020}. The English sentence \enquote{The doctor told the nurse that she had been busy.}, a human translator would resolve the co-reference of ‘she’ to the doctor, correctly translating it to `die {\"A}rztin'. However, a neural machine translation model (NMT) trained on a biased dataset, in which most doctors are male might incorrectly default to the masculine form, `der Arzt'. \cite{Hovy2020} investigated the prediction of age and gender of a text's author before and after translation. They found that \emph{Bing Translator}, \emph{DeepL Translator}, and \emph{Google Translate} all skew the predictions to be older and more masculine, which shows that NMT systems and the expression of gender interact.

According to Saunders et al. \cite{Saunders2020}, the first systematic analysis of gender bias in MT is from Stanovsky et al. \cite{Stanovsky2019}. They introduce \emph{WinoMT} which is the first challenge set explicitly designed to quantify gender bias in MT systems. Furthermore, they introduce an automatic evaluation method for eight different languages. Evaluating six MT systems, they found a strong preference for masculine translations in all eight gender-inflected languages. The above example sentence, where 'doctor' has been errouneousely translated into its male German form (\emph{Arzt}) is one of the $3,888$  sentences used in \emph{WinoMT}. 

\emph{WinoMT} focuses solely on the translation of occupations to evaluate gender bias. In this paper, we present the extended data set \emph{WiBeMT} to uncover gender bias in neural machine translation systems. Gender stereotypical occupations are augmented by gender stereotypical adjectives and verbs and we investigate how congruent and incongruent combinations impact translation accuracy of the NMT systems DeepL, Microsoft and Google Translate. In the next section, the original \emph{WinoMT} data set is introduced together with our extensions. Afterwards, results for the three NMT systems based on 70,686 sentences are presented, followed by a discussion and an outlook.

%% file: Methods.tex
\section{Constructing the Extended Challenge Set WiBeMT}

To construct a more diverse challenge set, we extend \emph{WinoMT}, respectively its core data base \emph{WinoBias}. Therefore, we identify verbs and adjectives of high stereotypicality with respect to gender. A gender score is determined by the cosine similarity of these words to a list of gender specific words. To calculate the similarity different pretrained word embeddings are used, where each of these words is represented by a vector. With the resulting most feminine and masculine adjectives the original sentences of \emph{WinoBias} are extended. Furthermore, new sentences are created combining occupations with gender stereotypical verbs.

\subsection{WinoBias and its Extension}
\emph{WinoMT} is based on two previous challenge sets, \emph{Winogender} \cite{Rudinger2018} and \emph{WinoBias} \cite{Zhao2018}. Both were introduced to quantify gender bias in co-reference resolution systems. Since \emph{WinoBias} constitutes $81.5\%$ of \emph{WinoMT}, we use it as basis for our extension. In total \emph{WinoBias} consists of $3,168$ sentences, of which $1,582$ are feminine and $1,586$ are masculine sentences. The sentences are based on $40$ occupations. An example sentence of \emph{WinoBias} involving a cleaner and a developer is: 
\begin{itemize}
    \item WinoBias sentence: \textit{The cleaner hates the \textbf{developer} because \textbf{she} always leaves the room dirty.}
    \item DeepL translation: \textit{Die Reinigungskraft hasst \textbf{den Entwickler}, weil \textbf{sie} das Zimmer immer schmutzig hinterlässt.}
\end{itemize}
\noindent
DeepL fails to correctly translate \textbf{developer} to the female inflection \textbf{die Entwicklerin}, but instead favors the stereotypical male inflection \textbf{der Entwickler}. 

The \emph{WinoBias} set is constructed such that each sentence is given in a stereotypical and an anti-stereotypical version. Stereotypical in this context means that the gender of the pronoun matches the predominant gender in the sentences occupation. The example sentence above is the anti-stereotypical version for the occupational noun 'developer', where the stereotypical version contains a 'he' instead of a 'she'. 

We argue that a measurement of gender bias solely based on the translation of occupational nouns does not do justice to the complexity of language. 
Therefore, we want to diversify the given approach by taking gender-stereotypical adjectives and verbs into account. This is realized in two steps:
First, \emph{WinoBias} sentences are extended such that each occupational noun is preceded with an adjective which is congruent or incongruent with respect to the gender stereotypical interpretation of the occupation. For instance, a developer might be \emph{eminent} (male) or \emph{brunette} (female). By adding feminine and masculine adjectives to each WinoBias sentence, we create a new subset of extended WinoBias sentences. 

Second, we create completely new sentences based on feminine and masculine verbs. One example with a feminine verb being: \enquote{The X \textbf{dances} in the club}, and one with a masculine verb: \enquote{The X \textbf{boasts} about the new car.}. Those base sentences are then extended with $99$ occupations from \cite{Zhao2018,Rudinger2018,Garg2018}, resulting in a new subset of verb sentences. All $100$ occupational nouns used in \emph{WinoBias} and in the new verb sentences are listed in Table~\ref{tab:WinoBiasOccupations2019}. 
After concatenating both subsets, the complete extended gender-bias challenge set consists of $70,686$ sentences. Overall, $100$ occupations are used in the set, $42$ gender-verbs, and $20$ gender-adjectives.

We hypothesize that gender-stereotypical verbs and adjectives influence the gender of the translation of the occupational nouns, when translating from English to the gender-inflected language German:

\vfill
\newpage
\begin{itemize}
    \item Hypothesis 1
    \begin{itemize}
        \item [] Sentences with a feminine verb result in significantly more translations into the female inflection of an occupation than sentences with a masculine verb.
    \end{itemize}
    \item Hypotheses 2
    \begin{itemize}
        \item [2a] WinoBias sentences extended with a feminine adjective result in significantly more translations into the female inflection of an occupation than original WinoBias sentences (without a preceded adjective).
        \item [2b] WinoBias sentences extended with a masculine adjective result in significantly more translations into the masculine inflection of an occupation than original WinoBias sentences (without a preceded adjective).
    \end{itemize}
\end{itemize}

\begin{table}[t]
	\centering
	\caption{All $100$ occupations, used in this thesis with the corresponding percentage of women in the occupation in the US.}
	\resizebox{0.9\textwidth}{!}{
	\begin{tabular}{lc|lc|lc|lc}
		\toprule
		Occupation & \% & Occupation & \% & Occupation & \% & Occupation & \% \\
		\midrule
		electrician & 2 & \textbf{analyst} & \textbf{41*} & \textbf{tailor} & \textbf{75} & dancer & -- \\
		\textbf{carpenter} & \textbf{3} & \textbf{physician} & \textbf{41} & \textbf{attendant} & \textbf{76*} & doctor & -- \\
		firefighter & 3 & surgeon & 41 & \textbf{counselor} & \textbf{76} & economist & -- \\
		plumber & 3 & \textbf{cook} & \textbf{42} & \textbf{teacher} & \textbf{78*} & educator & -- \\
		\makecell[l]{\textbf{construction-}\\\textbf{worker}} & \textbf{4} & chemist & 43 & planner & 79 & engineer & -- \\
		\textbf{laborer} & \textbf{4*} & \textbf{manager} & \textbf{43*} & \textbf{librarian} & \textbf{80} & examiner & -- \\
		\textbf{mechanic} & \textbf{4*} & \textbf{supervisor} & \textbf{44*} & psychologist & 80 & gardener & -- \\
		\textbf{driver} & \textbf{6*} & \textbf{salesperson} & \textbf{48*} & \textbf{assistant} & \textbf{85*} & geologist & -- \\
		machinist & 6 & photographer & 49 & \textbf{cleaner} & \textbf{89*} & hygienist & -- \\
		painter & 9 & bartender & 53 & \textbf{housekeeper} & \textbf{89*} & inspector & -- \\
		\textbf{mover} & \textbf{18*} & dispatcher & 53 & \textbf{receptionist} & \textbf{89} & investigator & -- \\
		\textbf{sheriff} & \textbf{18} & judge & 53 & \textbf{nurse} & \textbf{90*} & mathematician & -- \\
		\textbf{developer} & \textbf{20*} & artist & 54 & paralegal & 90 & officer & -- \\
		programmer & 20 & \textbf{designer} & \textbf{54} & dietitian & 92 & pathologist & -- \\
		\textbf{guard} & \textbf{22*} & \textbf{baker} & \textbf{60} & \textbf{hairdresser} & \textbf{92} & physicist & -- \\
		architect & 25 & pharmacist & 60 & nutritionist & 92 & practitioner & -- \\
		\textbf{farmer} & \textbf{25} & \textbf{accountant} & \textbf{62} & \textbf{secretary} & \textbf{93} & professor & -- \\
		\textbf{chief} & \textbf{28} & \textbf{auditor} & \textbf{62} & administrator & -- & sailor & -- \\
		dentist & 34 & \textbf{editor} & \textbf{63} & advisor & -- & scientist & -- \\
		paramedic & 34 & \textbf{writer} & \textbf{63*} & appraiser & -- & soldier & -- \\
		athlete & 35 & author & 64 & broker & -- & specialist & -- \\
		\textbf{lawyer} & \textbf{36} & instructor & 65 & CFO & -- & student & -- \\
		\textbf{janitor} & \textbf{37} & veterinarian & 68 & collector & -- & surveyor & -- \\
		musician & 37 & \textbf{cashier} & \textbf{71} & conductor & -- & technician & -- \\
		\textbf{CEO} & \textbf{39*} & \textbf{clerk} & \textbf{72*} & CTO & -- & therapist & -- \\
		\bottomrule
	\end{tabular}
	}
	\label{tab:WinoBiasOccupations2019}
\end{table}

\subsection{Finding Gender-Stereotypical Verbs and Adjectives}

To determine gender-stereotypicality of adjectives and verbs, large collections of these word types have been scored with respect to their similarity to a list of gender specific words given by \cite{Bolukbasi2016} As input we used a list of $3.250$ verbs from patternbasedwriting.com\footnote{\url{www.patternbasedwriting.com} offers teaching materials for primary school children.} and a combined list of $4.889$ adjectives from patternbasedwriting.com and from Garg et al. \cite{Garg2018}.

Calculation of gender score is based on word embeddings and the cosine-similarity, following the work of \cite{Bolukbasi2016,Garg2018}.
Their, similarity of the words to the pronouns \enquote{she} and \enquote{he} has been used. We extended scoring using a longer list of feminine and masculine words such as \enquote{mother}, \enquote{uncle}, \enquote{menopause} or \enquote{semen} to enhance robustness of the gender-score. This list, containing $95$ feminine and $108$ masculine words, is taken from \cite{Bolukbasi2016}, who used it for their debiasing methods of word embeddings.

Two families of \emph{word embeddings} were used: two pre-trained \emph{fastText}\footnote{Downloaded from: \url{https://fasttext.cc/docs/en/english-vectors.html}} from \cite{Mikolov2017} and two pre-trained versions of \emph{GloVe}\footnote{Downloaded from: \url{https://nlp.stanford.edu/projects/glove/}} from \cite{Pennington2014}. All four word embeddings have a vector size of 300 dimensions. Table \ref{tab:wordembeddingdataorigin} gives an overview of all word embeddings and their training data.

\begin{table}[t]
	\centering
	\caption{Summary of the origin of training corpora and their corresponding size for each word embedding.}
	\resizebox{0.9\textwidth}{!}{%
	\begin{tabular}{lcccc}
		\toprule
		& \multicolumn{4}{c}{Size [billion]} \\
		\cmidrule(r){2-5}
		corpora & \hspace{2pt}fastText-small\hspace{2pt}  &  \hspace{2pt}fastText-large\hspace{2pt} & \hspace{2pt}GloVe-small\hspace{2pt} & \hspace{2pt}GloVe-large\hspace{2pt} \\
		\midrule
		Wikipedia 2014 & --- & --- & 1.6 & --- \\
		
		Gigaword 5 & --- & --- & 4.3 & --- \\
		
		Wikipedia meta-page 2017 & 9.2 & --- & --- & --- \\
		
		Statmt.org News & 4.2 & --- & --- & --- \\
		
		UMBC News & 3.2 & --- & --- & --- \\
		
		Common Crawl & --- & 630 & --- & 840 \\
		\bottomrule
	\end{tabular}
    }
	\label{tab:wordembeddingdataorigin}
\end{table}

\emph{Cosine Similarity} is a measure of similarity between two normalized and non-zero vectors and can take values in the range between -1 and 1. Equation \ref{equ:CosineSimilarity} shows the calculation of the cosine similarity between the vectors $\mathbf{a}$ and $\mathbf{b}$ of two words:

\begin{equation}\label{equ:CosineSimilarity}
		\mathbf{similarity}=\cos (\theta)=\frac{\mathbf{a} \cdot \mathbf{b}}{\|\mathbf{a}\|\|\mathbf{b}\|}=\frac{\sum\limits_{i=1}^{n} a_{i} b_{i}}{\sqrt{\sum\limits_{i=1}^{n} a_{i}^{2}} \sqrt{\sum\limits_{i=1}^{n} b_{i}^{2}}}\\
\end{equation}

\noindent where $a_{i}$ and $b_{i}$ are components of vector $\mathbf{a}$ and $\mathbf{b}$ respectively.

Since word embeddings inherit to some extend the meaning of words, it is possible to use the cosine similarity as a measure for the similarity in meaning or, furthermore, the relationships between words. This enables mathematical operations on the vectors representing words such that: $cos(\overrightarrow{brunette} \cdot \overrightarrow{her}) \geq cos(\overrightarrow{brunette} \cdot \overrightarrow{him})$ becomes true for \enquote{brunette} and other gender-biased adjectives. If the feminine-gender value ($cos(\overrightarrow{brunette} \cdot \overrightarrow{her})$) is then subtracted from the masculine-gender value ($cos(\overrightarrow{brunette} \cdot \overrightarrow{him})$) the resulting single float value indicates whether a word is gender-biased in the word embedding with which the cosine-similarity was computed.

The total gender-score is the sum of eight single scores resulting from the combination of the four different word embeddings with the cosine similarity with \enquote{she} and \enquote{he} and with the list of feminine and masculine words. Since different word embeddings vary in the strength of the inherited gender-bias and all word embeddings should have equal impact on the overall score, the interim results were normalized to fit a range between $-1$ and $1$: 

\begin{equation}\label{equ:normalization}
	x^{\prime}=a+\frac{(x-\min (x))(b-a)}{\max (x)-\min (x)}
\end{equation}

\begin{figure}[t]
    \centering
	\includegraphics[width=0.9\linewidth]{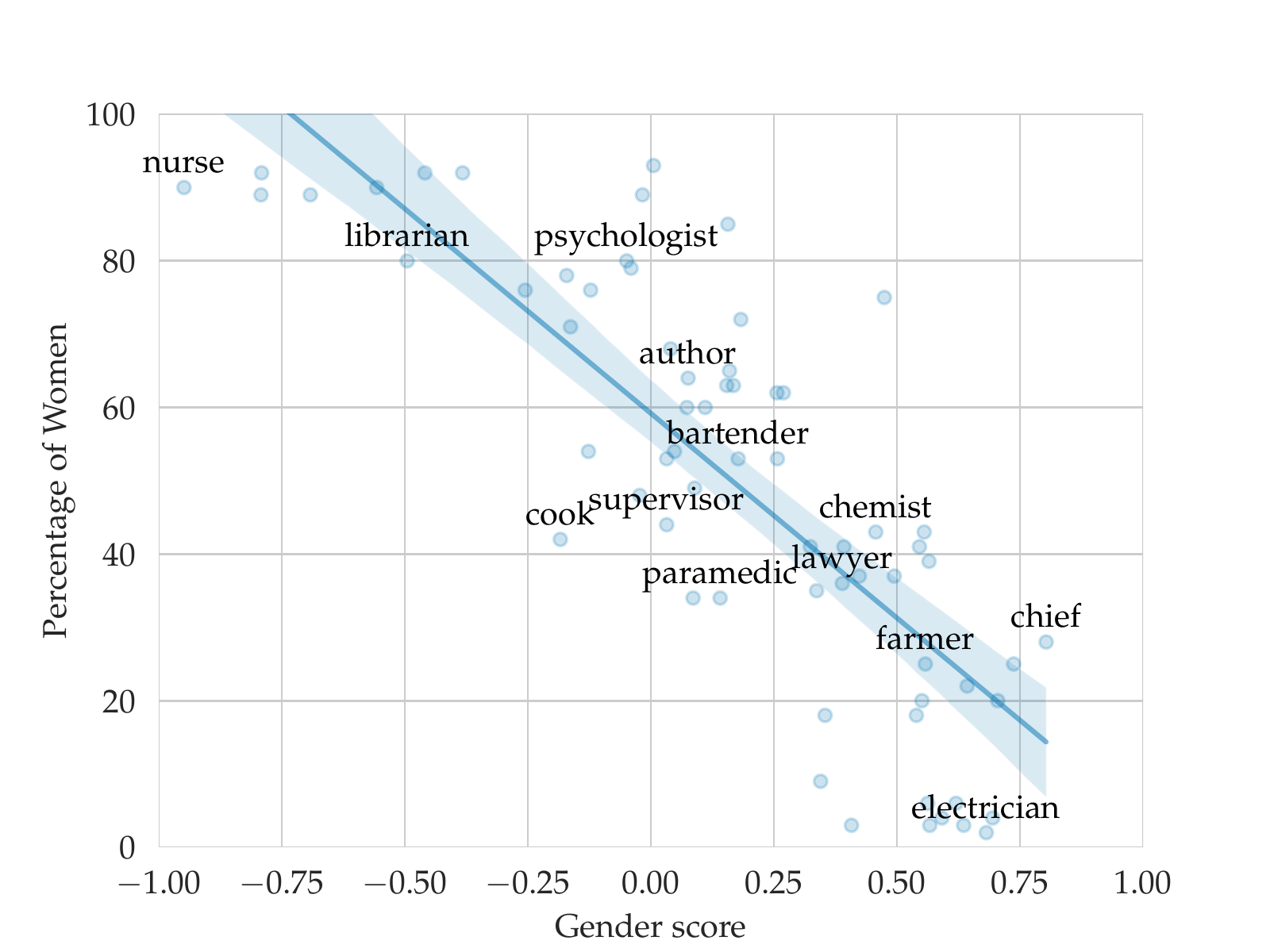}
	\caption{Scatter plot of the $99$ single word occupations with the percentage of wo\-men in the profession y-axis and the gender score on the x-axis.}
	\label{fig:occ_gender_score_percentage}
\end{figure}

To validate this procedure to determine a gender score for adjectives and verbs, the same method has been applied to the occupations given in Table~\ref{tab:WinoBiasOccupations2019}. The gender-score for these occupations shows a strong correlation to the percentage of women working in each given occupation (see Figure~\ref{fig:occ_gender_score_percentage}).

Gender-score has been calculated for all verbs and adjectives for which word embeddings existed. They were sorted ascending from the most negative -- and therefore most feminine -- score value, to the most positive, i.e., most masculine, value.

Of the $3,250$ \emph{verbs}, $3,210$ could be ranked ($med=0.151$, $std=0.165$)

After sorting the verbs by their gender-score, they were analyzed for their suitability to build a sentence in which person $P$ actively does action $A$.  The gender score of the $21$ selected feminine verbs ranges from $-0.772$ to $-0.233$ and of the masculine verbs from $0.733$ to $0.445$:

\begin{itemize}
    \item Feminine verbs: crochet, sew, accessorize, bake, embroider, primp, gossip, shriek, dance, undress, milk, giggle, marry, knit, twirl, wed, flirt, allure, shower, seduce, kiss;
    \item Masculine verbs: draft, tackle, swagger, trade, brawl, reckon, preach, sanction, build, boast, gamble, succeed, regard, retire, chuck, overthrow, rev, resign, apprehend, appoint, fool.
\end{itemize}
\paragraph{Adjectives: }
Of the $5,441$ \emph{adjectives}, $4,762$ could be ranked ($med=0.189$m $std=0.142$). 
After calculating the gender-score for the  $4,762$ adjectives, beginning with the most feminine, respectively, most masculine, each adjective was tested for its suitability considering the extension of existing WinoBias sentences. Words that semantically could not be combined with occupations were discarded (examples: hormonal, satin, luminous, philosophical, topographical). Furthermore, adjectives like \enquote{pregnant} -- which only apply to persons with a uterus -- were also discarded. The gender score of the $10$ selected feminine adjectives ranges from $-0.600$ to $-0.205$ and of the masculine adjectives from $0.654$ to $0.480$:

\begin{itemize}
    \item Feminine adjectives: sassy, perky, brunette, blonde, lovely, vivacious, saucy, bubbly, alluring, married;
    \item Masculine adjectives: grizzled, affable, jovial, suave, debonair, wiry, rascally, arrogant, shifty, eminent.
\end{itemize}

\subsection{Translation of \emph{WiBeMT} and Evaluation Design}
To test the extent to which machine translation systems inherit a gender-bias, three different services were tested with \emph{WiBeMT}: \emph{Google Translate}, \emph{Microsoft Translator}, and \emph{DeepL Translator}. All three NMT systems were accessed via their API, and all translation processes took place in July 2020.

To test our hypotheses that adjectives and verbs significantly influence the gender in translations of NMT systems, translations from English to German have to be categorized as either (correctly) feminine, masculine, neutral, or wrong. \cite{Stanovsky2019} use an automated method in the form of different morphological analyzers such as \emph{spaCy} to determine the gender of the occupation in the translated sentence. While this method is convenient for an automated approach that other researchers can use with different data, it also suffers from a certain degree of inaccuracy. To measure the accuracy of their automated evaluation method, \cite{Stanovsky2019} compared the automated evaluations with a random sample of samples evaluated by native speakers. They found an average agreement of 87\% between automated and native speaker evaluations.
To evaluate NMT systems with respect to \emph{WiBeMT}, we preferred to augment automated evaluation by \enquote{manual} evaluation to gain higher accuracy.

Evaluating the gender of all translations is based on a nested list, we refer to as \emph{classification-list}, and a set of rules for automated evaluation and manual evaluation for all remaining ambiguous translations. 
The classification-list contains four sublists for each occupation: a sublist of correct feminine translations, a sublist of correct masculine translations, a sublist of correct neutral translations, and a sublist of inconclusive or wrong translations. The gender of the translated occupation $Occ$ is then classified by checking in which sublist $Occ$ is listed and the gender is labeled as the sublist's category.

To control classifications for possible errors, $N=665$ (1\%) of the WinoBias sentences extended by adjectives were manually controlled for each translation system (in total $N=1,995$ sentences).
Not a single one was miscategorized. Of the verb-sentences translations, even 5\% were manually controlled by the authors, and here also, not a single one of the $618$ translations was miscategorized.

%% file: Results.tex
\section{Results}
After the creation of the \emph{WiBeMT} challenge set and the translation of all $70,686$ sentences with each NMT system, the translations were categorized as \emph{feminine}, \emph{masculine}, \emph{neutral}, or \emph{inconclusive / wrong}. The latter will be referred to as \enquote{wrong}. Due to these discrete categories the $Chi^{2}$ test of independence will be used for all statistical tests.
As the extended \emph{WinoBias} sentences include a pronoun that defines the gender, the results considering these data can be divided into \emph{true} and \emph{false}, and \emph{feminine} and \emph{masculine} translations. On the other hand, the verb sentences do not include any cue for a \emph{correct} gender in the translation. Therefore, they are just analyzed for feminine and masculine translations. The calculated \emph{feminine-ratio} ($\%TFG$) results from all feminine-translations divided by the sum of feminine- and masculine-translations. The calculated \emph{correct-gender-ratio} ($\%TCG$) results from all translations with correct gender divided by the sum of all translations with correct and incorrect gender. All $Chi^2$ were, if necessary, Bonferroni corrected. First, the results for the verb-sentences; Second, the results for the extended WinoBias sentences; And third, further results are presented.

\subsection{Hypothesis 1: Verb Sentences}
\begin{figure}[t]
	\centering
	\includegraphics[width=\linewidth]{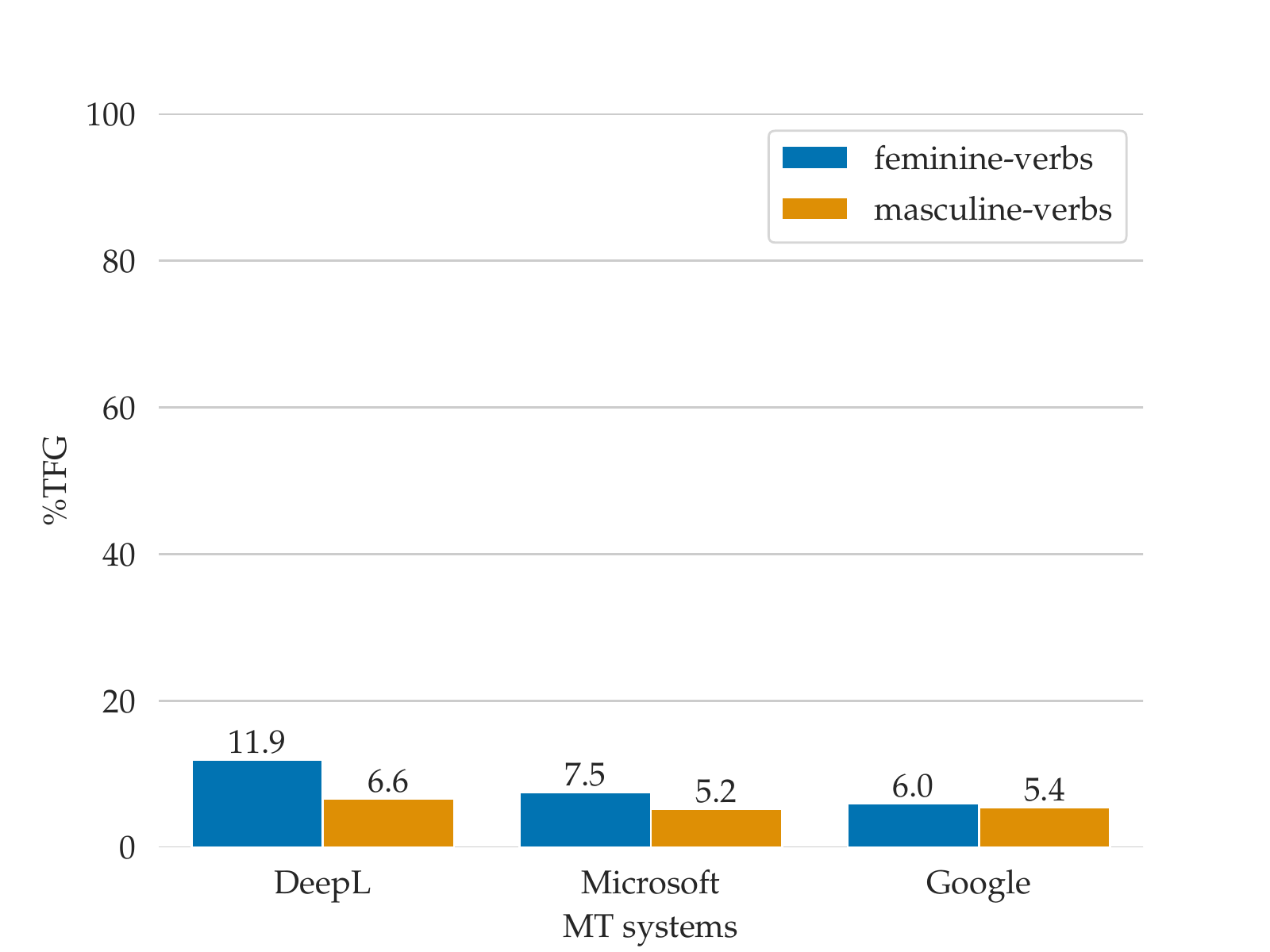}
	\caption{Percentage of translations to female gender ($\%TFG$) of occupations in verb sentences, organized by the category of verbs.}
	\label{fig:femaleRatioVerbs}
\end{figure}

The statistical tests regarding Hypothesis 1 yielded mixed results, depending on which NMT system is looked at. Of the $2.079$ sentences with a feminine verb DeepL translated the occupations in $242$ ($11.9\%$) sentences into the female gender ($\%TFG$) compared to Microsoft Translator with $149$ ($7.5\%$), and Google Translate with $120$ ($6.0\%$). Of the $2.079$ sentences with a masculine verb DeepL (DL) translated the occupations in $135$ ($6.6\%$) sentences into the female gender compared to Microsoft Translator (MS) with $104$ ($5.2\%$), and Google Translate (GT) with $109$ ($5.4\%$). For DeepL and Microsoft Translate the difference becomes significant with $\Delta\%TFG_{DL}=5.3\%$, $Chi^{2}_{DL}=33.7$ and $p_{DL}<0.001$, and $\Delta\%TFG_{MS}=2.3\%$, $Chi^{2}_{MS}=8.4$ and $p_{MS}=0.004$. For Google Translate the difference does not become significant. As two of three NMT systems inherit a gender bias that is influenced as expected by verbs, meaning that translations of sentences with feminine verbs result in a significantly higher $\%TFG$ than translations of sentences with masculine verbs, $H1$ is accepted. Figure \ref{fig:femaleRatioVerbs} gives an overview of the translations to the female gender ($\%TFG$) for all three NMT systems and the two categories of verb sentences.

\subsection{Hypothesis 2: Extended WinoBias Sentences}
Both hypotheses assume that adjectives that inherit a gender-bias in word embeddings also influence the gender of translations from English to German. While the results confirm the assumptions of Hypothesis H$2_{a}$, they also yield unexpected results for Hypothesis H$2_{b}$.
Of the 3,168 original WinoBias sentences DeepL translated $1,259$ ($41.7\%$), Microsoft Translator translated $1,041$ ($35.8\%$), and Google Translate translated $976$ ($33.2\%$) to the female inflection of the occupation. These percentages are used as the baseline to test, whether feminine and masculine adjectives skew the outcome of translations into the expected direction.

Of the sentences extended with a feminine adjective DeepL translated $49.5\%$, Microsoft Translator translated $42.5\%$, and Google Translate translated $40.1\%$ to the female inflection. The difference to the percentage of translations to the female inflection of the original WinoBias sentences becomes significant for all three NMT systems: $\Delta\%TFG_{DL}=7.8\%$, $Chi^{2}_{DL}=66.3$ and $p_{DL}<0.001$; $\Delta\%TFG_{MS}=6.7\%$, $Chi^{2}_{MS}=49.0$ and $p_{MS}<0.001$; and $\Delta\%TFG_{GT}=6.9\%$, $Chi^{2}_{GT}=53.8$ and $p_{GT}<0.001$. Therefore Hypothesis H$2_{a}$ is accepted.

\begin{figure}[t]
	\centering
	\includegraphics[width=\linewidth]{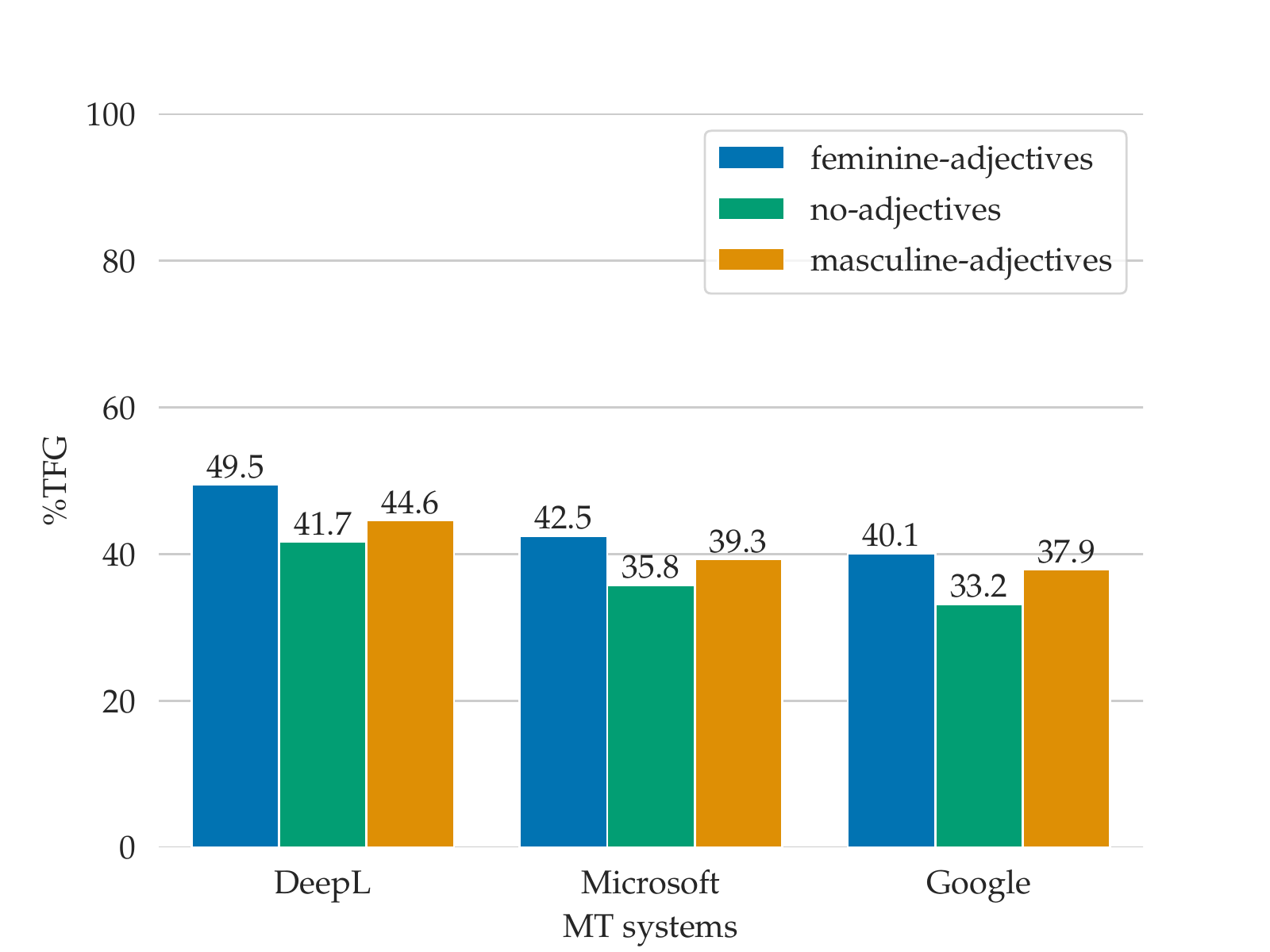}
	\caption{Percentage of translations to female gender ($\%TFG$) of occupations in all three types of EWB sentences: feminine adjective, masculine adjective, and no adjective.}
	\label{fig:EWB_Adjectives}
\end{figure}

Of the sentences extended with a feminine adjective DeepL translated $44.6\%$, Microsoft Translator translated $39.3\%$, and Google Translate translated $37.9\%$ to the female inflection. The difference to the percentage of translations to the female inflection of the original WinoBias sentences becomes significant for all three NMT systems: $\Delta\%TFG_{DL}=2.9\%$, $Chi^{2}_{DL}=9.1$ and $p_{DL}=0.008$; $\Delta\%TFG_{MS}=3.5\%$, $Chi^{2}_{MS}=13.3$ and $p_{MS}<0.001$; and $\Delta\%TFG_{GT}=4.7\%$, $Chi^{2}_{GT}=25.1$ and $p_{GT}<0.001$. While all differences are significant, they contradict our assumption that preceding masculine adjectives to WinoBias sentences results in less translations to the female inflection. Instead the preceded masculine adjectives have the opposite effect. Therefore, Hypothesis H$2_{b}$ is rejected.
Figure \ref{fig:EWB_Adjectives} gives an overview of the translations to the female gender ($\%TFG$) for all three NMT systems and the three categories of sentences. Table \ref{tab:EWBAdjectivesRawNumbers} lists all numbers of different translation categories for a deeper insight.

\begin{table}[t]
	\centering
	\caption{Numbers of the categorizations of the gender of translations of extended WinoBias sentences sorted by the gender-class of the adjective.}
	\begin{tabular}{lcccc}
		\toprule
		& & \multicolumn{3}{c}{NMT system} \\
		\cmidrule(r){3-5}
		adjective gender & translation category & DeepL & Microsoft & Google  \\
		\midrule
		feminine & 
		\makecell{true feminine \\ false masculine \\ true masculine \\ false feminine \\ neutral \\ wrong} & 
		\makecell{\textbf{13,341} \\ 1,553 \\ \textbf{13,833} \\ 1,751 \\ \textbf{550} \\ 652} & 
		\makecell{11,352 \\ 3,405 \\ 13,814 \\ 1,382 \\ 440 \\ 1,287} & 
		\makecell{10,177 \\ \textbf{4,593} \\ 13,312 \\ \textbf{1,826} \\ 472 \\ \textbf{1,300}} \\
		&&&&\\
		masculine & 
		\makecell{true feminine \\ false masculine \\ true masculine \\ false feminine \\ neutral \\ wrong} & 
		\makecell{\textbf{12,497} \\ 2,378 \\ \textbf{14,425} \\ 1,028 \\ \textbf{483} \\ 869} & 
		\makecell{8,966 \\ 3,405 \\ 11,943 \\ 963 \\ 346 \\ \textbf{6,057}} & 
		\makecell{8,724 \\ \textbf{4,533} \\ 12,128 \\ \textbf{1,450} \\ 447 \\ 4,398} \\
		&&&&\\
		no adjective & 
		\makecell{true feminine \\ false masculine \\ true masculine \\ false feminine \\ neutral \\ wrong} & 
		\makecell{\textbf{1,162} \\ 324 \\ \textbf{1,434} \\ 97 \\ 75 \\ 76} & 
		\makecell{916 \\ 503 \\ 1,365 \\ 125 \\ \textbf{77} \\ \textbf{182}} & 
		\makecell{812 \\ \textbf{652} \\ 1,313 \\ \textbf{164} \\ 53 \\ 174} \\
		\bottomrule
	\end{tabular}

	\label{tab:EWBAdjectivesRawNumbers}
\end{table}

\subsection{Influence of Gender-Stereotypical Verbs and Adjectives on Translations}
Our findings show that gender stereotypical verbs and adjectives influence gender bias in translations of NMT systems. In the following, we discuss our results in more detail.

\emph{Hypothesis 1:} In general, the results from verb sentences differ drastically from the ones of the extended WinoBias sentences, with far fewer sentences being translated to their feminine gender. The percentage of sentences translated to their feminine gender ($\%TFG$) in the verb sentences ranges from $5.2\%$ for masculine verb sentences translated by Microsoft Translator, to $11.9\%$ for feminine verb sentences translated by DeepL. In comparison to that, $\%TFG$ ranges from $33.2\%$ in original WinoBias sentences without adjective translated by Google Translate, to $49.5\%$ in extended WinoBias sentences with feminine adjectives translated by DeepL Translator.

Two reasons could be responsible for the low $\%TFG$. Firstly, the generic masculine in German: As mostly the masculine gender is used to address all genders, this must be present in the training data of all three NMT systems. Therefore, they tend to use the male inflection whenever the gender bias for a specific occupation does not outweigh the bias by the generic masculine. Secondly, the verb-sentences lack a gender pronoun like \enquote{her} or \enquote{him}, which urges the NMT system to decide which gender would be correct in the translation. This probably leads to the strong bias towards male inflections in translations.

To check whether the bias induced by occupations and the generic masculine outweighed an existing bias of verbs, we analyzed the data of the verb sentences again, looking at the different groups of occupations\footnote{With the calculated gender score we split the list of occupations in three equally sized categories (each $n=33$)}: feminine, neutral and masculine. For all three NMT systems the $\%TFG$ for verb sentences with neutral and masculine occupations was below 2\% regardless of the stereotypical feminine verbs. The $\%TFG$ in sentences with female occupations was 25.1\% in DeepL, 19.0\% in Microsoft Translator and 17.6\% in Google Translate. All differences to sentences with neutral and masculine occupations were significant with p-values below $0.001$ and $Chi^{2}$ values reaching from $352$ to $256$.

\emph{Hypotheses 2a \& 2b:} As gender bias works both ways in word embeddings, meaning that words can be stereotypical feminine or stereotypical masculine, we assumed that, depending on their gender-score derived from word embeddings, feminine adjectives would skew translations of NMT systems more often to their feminine gender and vice versa that masculine adjectives would skew translations more often to their masculine gender. This assumption was also supported by the findings of \cite{Stanovsky2019}, who preceded \enquote{handsome} to occupations of sentences which would be correct, if translated to their masculine gender, and \enquote{pretty} to occupations of sentences which would be correct if translated to their feminine gender. With this measure, they could improve the accuracy of NMT systems and reduce gender bias.

Contrary to the prior assumptions, adjectives preceded to occupations led to significantly more translations with feminine gender, regardless of their gender-score derived from word embeddings. Finding that masculine adjectives also lead to more translations with feminine gender does not only result in the rejection of H$2_{b}$, but also weakens the acceptance of H$2_{a}$. Therefore, we introduce a new Hypothesis H$2_{c}$: \enquote{WinoBias sentences extended with a feminine adjective result in significantly more translations into the female inflection of an occupation than WinoBias sentences extended with a masculine adjective.}. 

The difference between the percentage of translations to the female inflection of WinoBias sentences extended with feminine and WinoBias sentences extended with masculine adjectives becomes significant for all three NMT systems: $\Delta\%TFG_{DL}=4.9\%$, $Chi^{2}_{DL}=147.6$ and $p_{DL}<0.001$; $\Delta\%TFG_{MS}=3.2\%$, $Chi^{2}_{MS}=59.1$ and $p_{MS}<0.001$; and $\Delta\%TFG_{GT}=2.2\%$, $Chi^{2}_{GT}=29.2$ and $p_{GT}<0.001$. Therefore, Hypothesis H$2_{c}$ is accepted which strengthens H$2_{a}$, as stereotypical feminine adjectives preceded to occupational nouns lead to significantly more translations to the female gender than stereotypical masculine adjectives preceded to occupational nouns.


\subsection{Influence of Adjectives on Correct Gender in Translations}
The comparison of all three NMT systems was not one of the main research questions. Nevertheless, the extent of gender bias in the NMT systems can be of interest to users. Therefore, and because of the surprising findings considering Hypothesis $2_{b}$ a short comparison is presented in the following paragraphs.

\begin{figure}[t]
	\centering
	\includegraphics[width=\linewidth]{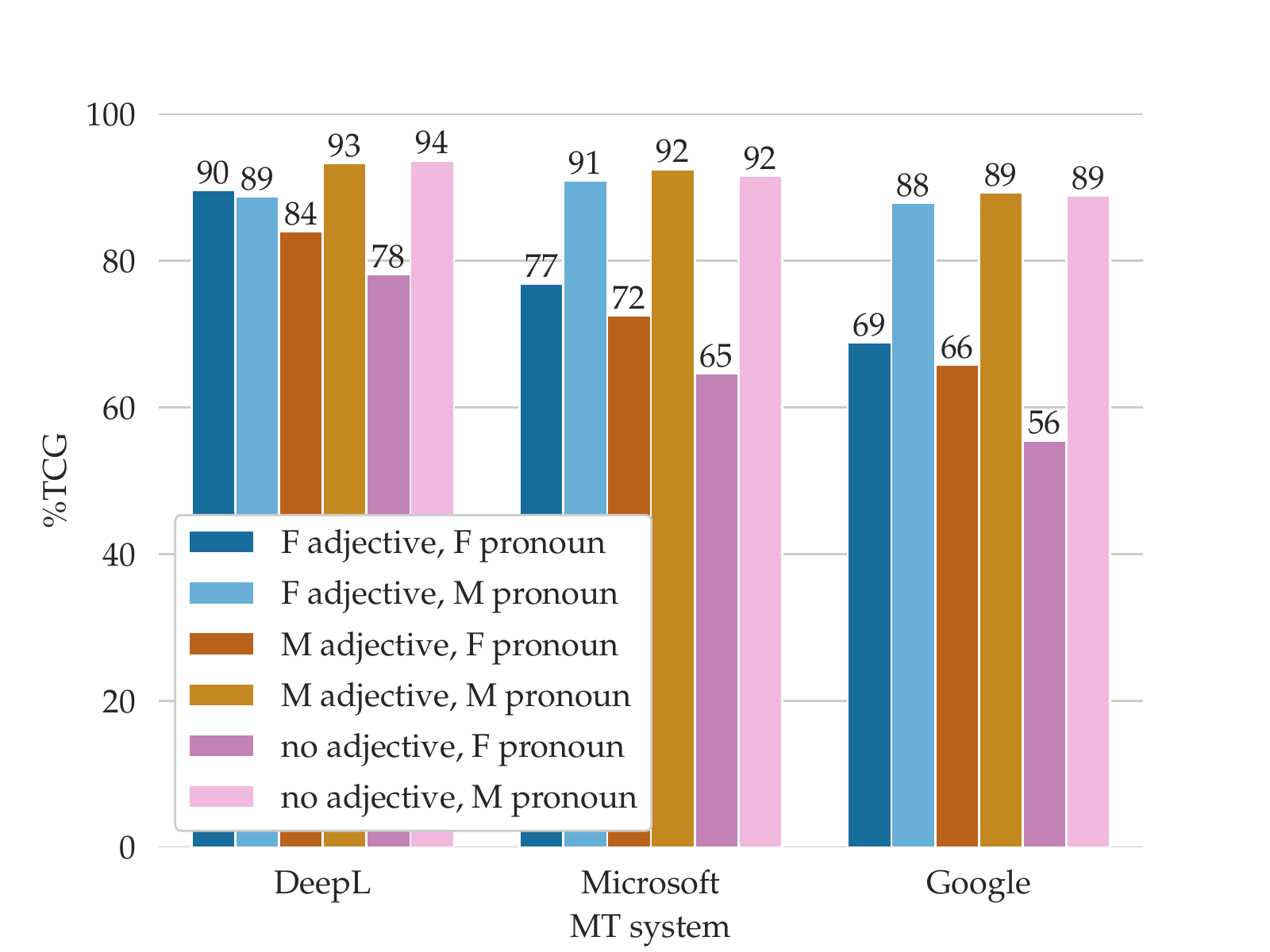}
	\caption{Bar plot of the percentage of translations into the correct gender ($\%TCG$) of original WinoBias sentences and extended WinoBias sentences.}
	\label{fig:BarplotEWBTCG}
\end{figure}

To better understand our results regarding hypotheses H$2_{a}$, H$2_{b}$ and H$2_{c}$ we plotted percentage of translations with the correct gender ($\%TCG$) organized by NMT systems and the type of sentence: original WinoBias sentence (no adjective), with masculine adjective extended WinoBias sentence (M adjective), and with feminine adjective extended WinoBias sentence (F adjective). Additionally, we split each type of sentences into sentences with male pronouns (M pronouns) and sentences with female pronouns (F pronouns). Figure \ref{fig:BarplotEWBTCG} shows the resulting plot. 

The results are quite astonishing: while the $\%TCG$ slightly decreases for sentences with male pronouns and any adjective ($1 \leq \Delta\%TCG \leq 5$) it drastically improves for sentences with female pronouns and any adjective ($12 \leq \Delta\%TCG \leq 13$). Considering that sentences with female and male pronouns are equally prevalent in original WinoBias sentences and extended WinoBias sentences, it shows that preceding an occupational noun with any adjective likely improves the overall percentage of translations to the correct gender inflection. This reflects the findings of \cite{Stanovsky2019}, who preceded \enquote{handsome} and \enquote{pretty} to occupational nouns of WinoMT sentences. With this measure, they could improve the accuracy hence $\%TCG$ of NMT systems and reduce gender bias. Our findings give more insight to this results: Stanovsky et al. \cite{Stanovsky2019} very likely could have added either \enquote{handsome} or \enquote{pretty} to all sentences, regardless of their pronoun and would nonetheless have been able to record a higher accuracy.

Furthermore, Figure \ref{fig:BarplotEWBTCG} shows that DeepL Translator performed best when it comes to $\%TCG$. Google Translate and Microsoft Translator again show more similar results, but Google Translate performed notably worse than Microsoft Translator. With the least discrepancy in $\%TCG$ between sentences with a feminine pronoun and sentences with a masculine pronoun in all three conditions (feminine adjective, masculine adjective, no adjective) DeepL Translator, therefore, inherits the lowest gender bias.

%% file: Discussion.tex
\section{Conclusions and Further Work}
The three neural machine translation systems evaluated with respect of their gender bias are black boxes, in so far as the architecture is not publicly available and -- even more important -- it is not transparent on what data these systems are trained. It is most likely that the gender bias in all three systems is inherited from the data used for training and their use of word embeddings.

To give a closer look on gender-bias of the NMT systems DeepL, Microsoft, and Google Translate, an extension of the \emph{WinoMT} challenge set -- 
the first challenge set designed to measure gender bias in NMT systems -- has been presented. While \emph{WinoMT} relies solely on the gender of the translation of occupations, our extended set \emph{WiBeMT} includes gender-adjectives and gender-verbs. Thereby, a more detailed assessment of gender bias has been possible. The number of sentences in our challenge set is, with over $70,000$ sentences, nearly $20$ times as large as the original WinoMT challenge set. This makes it less prone to overfitting when used to evaluate or reduce gender bias in NMT systems.

We could show that adjectives do significantly influence the gender of translated occupations. Against the hypotheses, feminine as well as masculine adjectives skew the translations of NMT systems to more translations with the feminine gender. Nevertheless, feminine adjectives still produce significantly more translations with the feminine gender than masculine adjectives do. 

All three NMT systems prefer translations to the generic masculine when no pronoun defines a correct gender for the translation. Despite this preference, $5.7\%$ to $9.3\%$ of all verb sentences are translated to their feminine gender by the MT systems. This can mostly be attributed to a gender bias in translating occupations, as our results regarding verb sentences and occupation categories show. The effect of gender-verbs on the gender of translations only became significant in DeepL and Microsoft Translator and was smaller than the effect of the occupations gender-categories. It can be assumed that Google Translate only performed better on the verb sentences task, as it generally has the strongest tendency to translate sentences to their masculine gender.

Surprisingly, gender-adjectives drastically improve the overall accuracy of all three NMT systems when it comes to the correct gender in translations. While the accuracy slightly drops for sentences with masculine pronouns, it drastically improves for sentences with feminine pronouns. Therefore, the discrepancy in accuracy between masculine and feminine pronoun sentences decreases, resulting in lower discrimination. One could even argue that gender-adjectives reduce gender bias in the output of NMT systems, but at the same time, it can be discriminating in itself that, as soon as you add an adjective to describe a person, the instance of this person is more likely to be translated to its feminine gender. Further research is certainly needed to find reasons for this effect and to assess the potentials of discrimination.

The sentences in our \emph{WiBeMT} challenge set are -- as the original \emph{WinoMT} challenge set -- constructed in a systematic way. While this allows for a controlled experiment environment, it might also introduce some artificial biases \cite{Stanovsky2019}. A solution could be to collect real-world examples of sentences, which are suitable for gender bias detection. Furthermore, the limitation on one source language (English) and one target language (German) does not allow for a generalization of the results. 

Another limitation of our study -- as most other studies -- is that it does not take into account that gender should not be seen as a binary, but rather a continuous variable. \cite{Cao2019}, for example, outline why \enquote{trans exclusionary} co-reference resolution systems can cause harm, which is probably also valid for MT systems. A further extension of the challenge set could help to shine a light on the shortcomings of the inclusion of transgender persons of NMT systems. 